\documentclass[conference]{IEEEtran}
\IEEEoverridecommandlockouts

\usepackage{amsmath,amssymb,amsfonts}
\usepackage{graphicx}
\usepackage[acronym]{glossaries}
\usepackage[hidelinks]{hyperref}
\usepackage{balance}
\usepackage{paralist}

\usepackage{enumitem} 

\newacronym{HRI}{HRI}{Human-Robot Interaction}

\begin{document}

\title{\LARGE \bf Common (good) practices measuring trust in HRI}

\author{
\IEEEauthorblockN{Patrick Holthaus}
\IEEEauthorblockA{\textit{Robotics Research Group}\\
University of Hertfordshire, Hatfield, United Kingdom.\\
\texttt{p.holthaus@herts.ac.uk}}
\and
\IEEEauthorblockN{Alessandra Rossi}
\IEEEauthorblockA{\textit{PRISCA Research Lab}\\
University of Naples Federico II, Naples, Italy.\\
\texttt{alessandra.rossi@unina.it}}
}
\maketitle

\section{Introduction}
Trust in robots is widely believed to be imperative for the adoption of robots into people's daily lives. It is, therefore, understandable that the literature of the last few decades focuses on measuring how much people trust robots -- and more generally, any agent - to foster such trust in these technologies.
Researchers have been exploring how people trust robot in different ways, such as measuring trust on human-robot interactions (HRI) based on textual descriptions or images without any physical contact~\cite{lee2021interactive,law2021interplay}, during and after interacting with the technology~\cite{Rossi2020,lingg2023building}.

Nevertheless, trust is a complex behaviour, and it is affected and depends on several factors, including those related to the interacting agents (e.g. humans, robots, pets), itself (e.g. capabilities, reliability), the context (e.g. task), and the environment (e.g. public spaces vs private spaces vs working spaces)\cite{Hancock2011,Rossi2017_2}. In general, most roboticists agree that insufficient levels of trust lead to a risk of disengagement~\cite{salem2015towards} while over-trust in technology can cause over-reliance and inherit dangers~\cite{aroyo2021overtrusting}, for example, in emergency situations~\cite{robinette2016overtrust}.
It is, therefore, very important that the research community has access to reliable methods to measure people's trust in robots and technology. In this position paper, we outline current methods and their strengths, identify (some) weakly covered aspects and discuss the potential for covering a more comprehensive amount of factors influencing trust in HRI.

\section{Current measures of trust}

Current trust measures often aim to quantify the amount of trust that is subconsciously exhibited towards a robot. Subjective questionnaire scales, such as ~\cite{ullman2019measuring,schaefer2016measuring,jian2000foundations}, consist of several items, which when combined can capture different aspects of trust. Individual aspects can, for example, reflect people's trust in performance or morality~\cite{ullman2019measuring}. Such questionnaire scales are widely adopted, but sometimes concerns regarding validity or positive bias emerge~\cite{gutzwiller2019positive}.
To account for such biases, subjective measurements are often complemented by objective measurements of trust. Researchers, for example, successfully use compliance with a robot's suggestions~\cite{salem2015would} or record whether people are sharing information that makes them potentially vulnerable -- such as secrets~\cite{Rossi2020} -- with a robot as indicators of whether people trust a robot. These choices strongly relate trust to the task that the robot is undertaking, such as its criticality of the task, whether it implies a cognitive or a physical task, and the magnitude of its consequences. 
Trust is usually measured within a specific scenario where the human-robot interaction takes place, sometimes also asking people to relate to other environments by imagining a different scenario or environment \cite{10.1007/978-3-030-90525-5_11,Rossi2020}, for example, by questioning whether their responses would change in another situation, with another agent, in a public vs private environment. 

To calibrate subjective and objective measurements of subconsciously exhibited trust and to account for potential external biases and situational influences, trust measures are often combined with other questionnaires that measure people's general tendency or propensity to trust~\cite{alarcon2018effect}, or their potentially negative attitude towards robots~\cite{nomura2004psychology}. Such questionnaires are usually administered before any interaction to avoid side effects of the experiment itself. When calibrating against or comparing individual robots, trust measures are also often combined with scales that are able to measure the social attributes of a specific robot (e.g.~\cite{carpinella2017robotic}). Moreover, to account for potential influences of novelty, researchers employ familiarisation phases where participants can get accustomed to the robot~\cite{edmonds2019tale} to minimise such an effect, or look into repeated and/or long-term experiments to directly observe any changes (e.g.~\cite{Rossi2020,Ayub2023c}).

\section{Drawbacks and opportunities}
The number of different approaches to measuring trust shows that there is currently no catch-all method providing a solution that would be viable across different experimental setups; individual solutions are required that are attuned to the participants, situation and research questions. Henceforth, a mix of methods is currently regarded as the gold standard. Appropriate questionnaire scales measuring more universal aspects of trust are usually paired with individual questions targeted at identifying further influences that emerge from the specific experimental situation. If employed, additional open-ended questions about people's choices or behaviours can complete that mix of methods and allow for further qualitative insights to support reasoning on top of the quantitative analysis.

However, even such a mix of methods cannot comprehensively account for the magnitude of factors that potentially affect the trust people put into an interaction with a robot. Firstly, we would like to mention that most quantitative analyses only measure trust subconsciously, and researchers often avoid direct questions about how much people trust a robot. Since we believe that the trust people put forward might also be related to conscious concepts of a robot's (social) credibility~\cite{Holthaus2021c}, we would like to encourage researchers to incorporate direct questions about a robot's trustworthiness and reasoning why people would trust or not trust a robot into their experimental scenario. Open-ended questions can also help to capture conscious aspects of trust when asking to explain if a person's behaviours were influenced by their trust in the robot.

In line with \cite{chita2021can}, we propose to engage in a discussion to complement existing trust measures by looking at the wider situation and work together to find new \textit{ad hoc} measures to address common side effects that might influence the amount of trust people put into a robot. Such measures would aim to assess:
\begin{inparaenum}[(I)]
    \item the impact of repeated interactions and long-term human-robot interactions \cite{Rossi2020} with particular attention to the novelty effect and how it might interplay with trust metrics;
    \item how imperfect interactions \cite{Schulz2021} might overshadow results;
    \item how subconscious interpretations \cite{Holthaus2023} might have been formed, and, as a consequence, influence the perception of behaviours;
    \item how the conformity of robot behaviours to social norms \cite{Holthaus2019} in accordance with the situation might affect findings; and 
    \item how the surroundings or experimental environment prevent or facilitate feelings of trust.
\end{inparaenum}
We, therefore, suggest specifically looking into five different aspects as a result of the current research that we believe are key for developing future measures of trust in HRI. These key aspects are the following:
\begin{enumerate}[labelindent = \parindent, leftmargin=2.1cm,labelsep=0.2cm]
    \item[Human] This aspect includes all factors related to the human, such as personality, predisposition to trust, previous experience and expectations.
    \item[Robot] We suggest measuring aspects related to the robot itself, such as people's perception of its appearance, its functionalities, and its performance.
    \item[Task] This aspect includes factors depending on the task's characteristics. For example, it would include concepts like outcome, severity and criticality.
    \item[Environment] The environment aspect aims to measure factors depending on the environment characteristics where the HRI is developing and how it may affect trust. These could include public, private and working environments. 
    \item[Other agents] This aspect could include factors depending on the presence and actions of other agents, such as pets, children, and older people.
\end{enumerate}

\section{Conclusion}
Trust is a fundamental aspect that drives not only human interactions but any human-agent interaction in the humans' day-to-day activities. Yet, this is one of the topics in HRI that still does not have well-defined and \textit{ad hoc} measures to evaluate to which extent people trust robots, and, as a consequence, to calibrate such trust to have a successful and effective relationship. Accordingly, this paper suggests investigating five key factors that affect trust within a situation to be able to assess and reduce common side effects influencing how people put their trust in robots.

\section*{Acknowledgments}
This work has been supported by the Italian PON R\&I 2014-2020 - REACT-EU (CUP E65F21002920003).

\balance

\bibliographystyle{IEEEtran}
\bibliography{bib}

\end{document}